\documentclass{article}
\include{graphicx} 

\newenvironment{scriptsizedescription}
   {\begin{description}\scriptsize}{\end{description}}
\newenvironment{scriptsizeenumerate}
   {\begin{enumerate}\scriptsize}{\end{enumerate}}
\newenvironment{scriptsizeitemize}
   {\begin{itemize}\scriptsize}{\end{itemize}}
\newenvironment{scriptsizeflushleft}
   {\begin{flushleft}\scriptsize}{\end{flushleft}}

\newcommand{\type}{ \colon }
\newcommand{\morph}{ \colon }
\newcommand{\memberof}{\,{\in}\,}
\newcommand{\define}{\stackrel{{\rm df}}{=}}
\newcommand{\tensor}{{\otimes}}
\newcommand{\implication}{{\Rightarrow}}

\newcommand{\minus}{\: \dot{-} \:}
\newcommand{\tensorimplysource}
   {\! \mbox{\hspace{.21em}--\hspace{-.21em}} \backslash }

\newcommand{\product}[2]{ {#1} {\times} {#2} }
\newcommand{\triproduct}[3]{ {#1} {\times} {#2} {\times} {#3} }

\newcommand{\pair}[2]{\langle #1,#2 \rangle}
\newcommand{\triple}[3]{\mbox{$ \langle #1,#2,#3 \rangle $}}
\newcommand{\quadruple}[4]{\mbox{$ \langle #1,#2,#3,#4 \rangle $}}
\newcommand{\quintuple}[5]{\mbox{$ \langle #1,#2,#3,#4,#5 \rangle $}}

\newcommand{\puttext}[3]{\put(#1,#2){{\mbox{\tiny$#3$\normalsize}}}}
\newcommand{\putdisk}[3]{\put(#1,#2){\circle*{#3}}}

\title{Soft Concept Analysis}
\author{Robert E.\ Kent}
\date{}

\begin{document}
	\maketitle

\section{Overview}

In this chapter we discuss soft concept analysis\cite{kent93,kent94},
a study which identifies 
an enriched notion of \emph{conceptual scale} as developed in formal concept  analysis
\cite{ganter89}
with an enriched notion of \emph{linguistic variable} as discussed in fuzzy logic 
\cite{zadeh75}.
The identification
\emph{enriched conceptual scale} $\equiv$ \emph{enriched linguistic variable}
was made in a previous paper \cite{kent94}.
In this chapter we offer further arguments for the importance of this identification
by discussing the philosophy, spirit, and practical application of conceptual scaling
to the discovery, conceptual analysis, interpretation, and categorization
of networked information resources.
We argue that a linguistic variable,
which has been defined at just the right generalization of valuated categories \cite{lawvere73},
provides a natural definition for the process of soft conceptual scaling.
This enrichment using valuated categories 
models the relation of indiscernability \cite{kent93,pagliani96,pagliani97},
a notion of central importance in rough set theory \cite{pawlak82}.
At a more fundamental level for soft concept analysis,
it also models the derivation of formal concepts\cite{kent92},
a process of central importance in formal concept analysis \cite{wille82}.
Soft concept analysis is synonymous with enriched concept analysis.
From one viewpoint,
the study of soft concept analysis that is initiated here
extends formal concept analysis to soft computational structures.
From another viewpoint,
soft concept analysis provides a natural foundation for soft computation
by unifying and explaining notions from soft computation
in terms of suitably generalized notions from formal concept analysis, rough set theory and fuzzy set theory.


\section{Networked Information Resources}

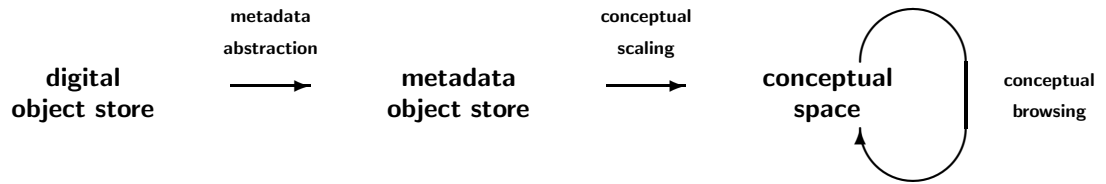
\begin{figure}
\begin{tabular}{cccccc}
\begin{tabular}{c}
	\normalsize{\sffamily\bfseries digital} \\
	\normalsize{\sffamily\bfseries object store}
\end{tabular}
&
\begin{tabular}{c}
	\scriptsize{\sffamily\bfseries metadata} \\
	\scriptsize{\sffamily\bfseries abstraction} \\
\begin{picture}(40,0)(0,0)\thicklines\put(5,0){\vector(1,0){30}}\end{picture} \\
	{\sffamily} \\
	{\sffamily} \\
	{\sffamily}
\end{tabular}
&
\begin{tabular}{c}
	\normalsize{\sffamily\bfseries metadata} \\
	\normalsize{\sffamily\bfseries object store}
\end{tabular}
&
\begin{tabular}{c}
	\scriptsize{\sffamily\bfseries conceptual} \\
	\scriptsize{\sffamily\bfseries scaling} \\
\begin{picture}(40,0)(0,0)\thicklines\put(5,0){\vector(1,0){30}}\end{picture} \\
	{\sffamily} \\
	{\sffamily} \\
	{\sffamily}
\end{tabular}
&
\begin{tabular}{c}
	\normalsize{\sffamily\bfseries conceptual} \\
	\normalsize{\sffamily\bfseries space}
\end{tabular}
&
\begin{picture}(15,0)(10,0)\thicklines
\put(0,15){\oval(40,40)[t]}
\put(20,15){\line(0,-1){25}}
\put(0,-10){\oval(40,40)[b]}
\put(-20,-10){\vector(0,1){0}}
\end{picture}
\begin{tabular}{c}
	\scriptsize{\sffamily\bfseries conceptual} \\
	\scriptsize{\sffamily\bfseries browsing} \\
\end{tabular}
\end{tabular}
\caption{{\bf The {\sffamily\bfseries WAVE} system architecture}}
\label{architecture}
\end{figure}

An information management software system for the World-Wide Web
called {\sffamily WAVE (Web Analysis and Visualization Environment)}
\footnote{Accessible at the Web address 
\texttt{http://wave.eecs.wsu.edu/}
.}
is currently under development \cite{kent:neuss:94}.
{\sffamily WAVE} is a third generation World-Wide Web tool 
used for conceptual navigation and discovery over a universe of networked information resources.
Figure~\ref{architecture} is a diagram of the architecture of the {\sffamily WAVE} system.
This consists of three major components
(the digital object store, the metadata object store, and the conceptual space),
and three processes (metadata abstraction, conceptual scaling, conceptual browsing) which connect the components.
The digital object store represents the information space for a community on the World-Wide Web
as stored in various Web document collections or online databases.
The metadata object store represents information abstracted from the digital object store.
The process of metadata abstraction includes extraction from raw HTML\footnote{HyperText Markup Language, the current lingua franca of the World-Wide Web.} documents or translation from annotated XML\footnote{eXtensible Markup Language, a data format for structured document interchange on the World-Wide Web; XML is a metalanguage used to define markup languages, which are called XML applications; see 
\texttt{http://www.w3.org/XML/}
.} files,
both done by Web robots.
The {\sffamily WAVE} system represents its metadata in a markup language (an XML application) called Ontology Markup Language (OML)\footnote{See 
\texttt{http://wave.eecs.wsu.edu/WAVE/Ontologies/OML/OML-DTD.html}
; Ontology Markup Language (OML) owes much to pioneering efforts of the SHOE initiative
(see 
\texttt{http://www.cs.umd.edu/projects/plus/SHOE/}
)
at the University of Maryland at College Park.}.
OML is a semantic data model
--- an extended form of the entity-relationship model of database theory.
OML represents information in terms of abstract objects and relations between those objects.
The type structure of this information is specified in OML by an ontology,
which consists of a generalization-specialization hierarchy or taxonomy of categories (object types)
and relational schemata (relation types) between those categories.

The conceptual space is a form of conceptual knowledge representation of the original information.
It organizes the ontological information of the metadata object store in terms of the formal concepts of formal concept analysis.
The conceptual space represents information conceptually scaled from the metadata object store.
The process of conceptual scaling applies conceptual scales to the metadata captured in the ontological categories and relations of OML,
thereby transforming it into a framework suitable for the concept lattices of formal concept analysis\cite{wille82}.
An extension of OML called Conceptual Knowledge Markup Language (CKML)\footnote{See 
\texttt{http://wave.eecs.wsu.edu/WAVE/Ontologies/CKML/CKML-DTD.html}
; CKML follows the philosophy and practice of conceptual knowledge processing, a principled approach to knowledge representation and data analysis\cite{wille82}.}
specifies in the {\sffamily WAVE} system the conceptual scales used in the scaling process.
The conceptual space component is essentially a concept lattice with a naming facility for bookmarking favorite formal concepts.
Bookmarked concepts are called conceptual views.

The {\sffamily WAVE} system has a client/server architecture.
The metadata object store component,
the conceptual scaling process,
and the conceptual space construction,
all reside or take place on the {\sffamily WAVE} server.
The {\sffamily WAVE} system has an online client software interface
that allows the user to download the conceptual space and browse over its lattice of formal concepts.
The {\sffamily WAVE} system is a state transition system
with formal concepts playing the role of states 
and the concept lattice functioning as a state space.
The conceptual browsing process provides state transitioning over the concept lattice
by either of two methods:
direct transition to named concepts by mouse click;
and
definition of next conceptual state by use of lattice meets and joins.
The process of conceptual browsing is dual mode: extensional and intentional.
The extensional browsing mode,
which visibly browses globally over conceptual views and attribute generated concepts,
defines the next conceptual state in terms of the meets and joins of a collection of views restricted to the meets of a collection of attributes.
Neighboring concepts, neither above or below the current concept,
are compared to the current conceptual state in terms of the objects common to their extents.
The intentional browsing mode is dual.


\begin{figure}

\includegraphics[height=2.5in,width=4.25in]{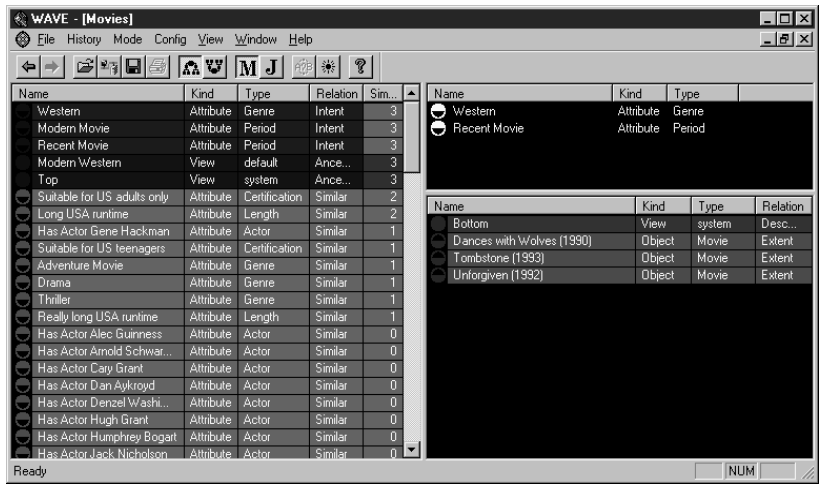}

\caption{{\bf The {\sffamily WAVE} conceptual interface}}
\label{interface}
\end{figure}


\begin{center}
\fbox{\begin{minipage}{11cm}\scriptsize

Figure~\ref{interface} is an image of the {\sffamily WAVE} system client interface 
taken during conceptual browsing over a movie information space.
The extensional browsing mode is being used here
(as indicated by the depressed inverted tree symbol button).
The three panes in Figure~\ref{interface} are:
the definition pane on the upper right, 
which lists elements that are being used in the definition of the current concept;
the global pane on the left, 
which is a global display of all view and attribute concepts in the conceptual space;
and
the local pane on the bottom right, 
which is a local display of both the extent of the current concept and views more specific (below) the current concept.
Named formal concepts are visible as explicit entries in the client interface:
objects, attributes and conceptual views.
The kind column indicates whether the lattice element is an object, attribute or view.
The type column lists the scope category for an object, the conceptual scale of an attribute, and the defining/owning agent for a conceptual view.
The relation column
gives an ``Equivalent'' label for a lattice element which labels the current conceptual state,
gives an ``Intent'' (``Extent'') label for an attribute (object) which in the intent (extent) of the current state,
gives an ``Ancestor'' (``Descendant'') label for a view which above (below) the current state in the lattice order,
and
shows a ``Similar'' label for a lattice element which is off to the side
--- neither above nor below the current concept.
Finally,
the similarity column displays the extensional similarity 
between a lattice element and the current conceptual state;
in extensional mode elements above the current state have maximal similarity.
The current conceptual state in Figure~\ref{interface} is 
the lattice meet 
(the depressed big M symbol button)
of 
(1) the ``Western'' attribute defined by a nominal conceptual scaling of 
a composite description function consisting of just the binary relation
``genre'' between movie ontology categories ``Movie'' and ``Genre''
(``Western'' is defined in CKML by the query ``What movies have western genre?'')
and 
(2) the ``Recent Movie'' attribute defined by an ordinal conceptual scaling of 
the simple description function ``year'' from the category ``Movie'' to the datatype ``Date''
(``Recent Movie'' is defined in CKML by the query ``What movies appeared in year $\geq 1990$?'').
The current concept in Figure~\ref{interface} does not have a name;
that is,
it is not labeled as a conceptual view.
If the current concept was a view, 
that view name would appear in the global pane with an ``Equivalent'' value in the relation column.
Anonymous concepts such as this can be given a name by using them to define a new conceptual view
(the bright dot symbol button)
--- conceptual views are named formal concepts.
The extent of the current conceptual state in Figure~\ref{interface} is the collection of three movies listed in the local pane.
The similarity column in the global pane shows that the current concept has one object in common with the attribute ``Suitable for US teenagers''.
Extensional similarity,
which provides the user with a sideways neighboring dimension within the concept lattice,
is defined as the extent cardinality of the lattice meet of two concepts.

\normalsize\end{minipage}}
\end{center}

The analogies in Table~\ref{analogies} 
link the various major components and processes in the architecture of {\sffamily WAVE} with the traditional library,
and illustrate the fact that the {\sffamily WAVE} system is a digital library.
Interpretation of resource descriptions,
via conceptual scaling or faceted analysis,
plays a central role in the {\sffamily WAVE} system.
At the present time,
the {\sffamily WAVE} system conceptually analyzes, interprets, and categorizes resources,
such as Web textual and image documents,
in a crisp or hard fashion.
However,
using ideas developed in this chapter,
an excellent approach for the extension to an enriched {\sffamily WAVE} system
is quite clear.
The following short list of conceptually scalable attributes
indicates that notions of approximation are very important 
for networked information resources:
the visible size of textual documents in pages or some other meaningful unit;
the concept extent cardinality as a count of equivalent instances of resources;
similarity measures between Web documents based upon numbers of common attributes;
relative scores for search engine keyword search;
the cost of resources;
the duration of play for audio/video data;
the critical review of resources;
etc.
As a particular example of some importance,
the co-occurrence matrices of semantic retrieval\cite{chen:etal94} 
effectively represent networked information as weighted or fuzzy formal contexts. 
An enriched {\sffamily WAVE} system
will allow the user to define according to his own judgement
various enriched interpretations of networked resource information.

\begin{table}
\fbox{\scriptsize\begin{tabular}{|c|c|}\hline
{\sffamily\bfseries The Traditional Library}
&
{\sffamily\bfseries The WAVE System} 
\\ \hline\hline 
{\sffamily\bfseries holdings} 
&
{\sffamily\bfseries digital object store}
\\ \hline 
{\sffamily\bfseries catalogs and indexes}
&
{\sffamily\bfseries metadata object store}
\\ \hline 
\begin{tabular}{c}
{\sffamily\bfseries classification scheme} \\ \hline
Dewey, LC, Colon, etc.
\end{tabular}
&
{\sffamily\bfseries concept space} 
\\ \hline 
{\sffamily\bfseries cataloging practices}
&
\begin{tabular}{c}
{\sffamily\bfseries metadata abstraction}
\\
{\sffamily\bfseries conceptual scaling}
\end{tabular}
\\ \hline 
{\sffamily\bfseries reference librarian}
&
{\sffamily\bfseries conceptual browsing} 
\\ \hline
\end{tabular}}
\caption{{\bf Analogies}}
\label{analogies}
\end{table}


\section{The Process of Conceptual Scaling}


The application of facets in the theory of library classification
was first tested and developed by Ranganathan in his Colon classification system.
Faceted analysis and classification provides a flexible means to classify
complex, multi-concept subjects\cite{rowley87}.
Complex subjects are divided into their component, single-concept subjects.
Single-concept subjects are called {\em isolates\/}.
Faceted analysis examines the literature of an area of knowledge and identifies its isolates.
A {\em facet\/} is the sum total of isolates formed by the division of a subject by one characteristic of division.
Some examples of facets in musical literature are:
composer, instrument, form, etc.
Isolates within facets are known as foci.

\begin{table}
\fbox{\scriptsize\begin{tabular}{|c|c|c|}\hline
	{\bf faceted analysis} 	& {\bf conceptual scaling}  	& {\bf linguistic variable use} \\ \hline\hline
	facet           	& conceptual scale 		& linguistic variable \\
	isolate/foci		& scale attribute		& linguistic value \\ \hline
\end{tabular}}
\caption{{\bf Identities}}
\label{identities}
\end{table}

Comparing these ideas to ontologically structured metadata\cite{kent94},
facets are identified with conceptual scales or linguistic variables
and are often associated with a composite description function,
and isolate/foci are identified with scale attributes or linguistic values.
A composite description function in the ontology
may consist of one function with primitive image values,
or a binary relation connecting two categories of objects composable with such a function,
or something more complex.
These observations are recorded in Table~\ref{identities}.
In conceptual knowledge processing each facet is computed by a conceptual scale.
A conceptual scale is an active filter or lens
through which information is interpreted.
Faceted analysis is conceptual scaling.
It involves four steps.
\begin{scriptsizeenumerate}
	\item Gather ontologically structured metadata.
	\item Identify conceptual scales of interest and specify attributes within them.
	\item Specify the structure of conceptual scales.
	\item Apply the conceptual scales to the metadata,
		producing a composable vector of facets which constitutes the conceptual space.
\end{scriptsizeenumerate}


As stated in the following principles of interpretation and classification,
our use of information involves our interpretion of it.
\begin{scriptsizeitemize}
	\item {\bf Information use is interpretation:}
		Humans use information by representing it in conceptual structures.
		Such conceptual structures are constructable.
		This construction is partly interpretive and partly automatic.
		The design of the conceptual constructors,
		although governed by principles,
		requires interactive advice from human experts.
		The application of the conceptual constructors,
		the actual categorization of information 
		and the construction of conceptual classes,
		can be automatic.
	\item {\bf Interpretation involves classification:}
		Interpretation defines (implicit) conceptual structures.
		Since explicit categories are special conceptual structures
		(tree structures for single inheritance or directed acyclic graphs for multi-inheritance),
		specified classification can help define interpretation.
	\item {\bf Conceptual classification is general:}
		Most conceptual classification structures are hierarchical tree structures.
		But tree structures are special cases of lattice structures
		--- just add a bottom node
		which is the meet for all pairs of categories;
		the join of two categories is their most specific ancestor.
	\item {\bf Conceptual classification is composable:}
		Conceptual classification schemes can be composed,
		using summing operations (such as apposition)
		and/or producting operations,
		from a small set of primitive conceptual structures called {\em conceptual scales\/}.
	\item {\bf Conceptual generation is polar:}
		Each collection of instances (objects) generates a category
		(conceptual class).
		The smaller the collection of objects,
		the more prototypical and exemplary they are of the category that they generate.
\end{scriptsizeitemize}

\begin{table}
\fbox{\scriptsize\begin{tabular}{|c|c|c|}\hline
	{\bf scale type}   & {\bf mathematical structure} & {\bf intuitive idea} \\ \hline\hline
	{\bf nominal}      & set & partition/separateness \\
	{\bf ordinal}      & (often) total order & ranking \\
	{\bf interordinal} & partial order of intervals & betweenness  \\
	{\bf hierarchical} & tree structure & nesting \\
	{\bf metrical}     & generalized metric space & similarity \\ \hline
\end{tabular}}
\caption{{\bf Conceptual scales}}
\label{scales}
\end{table}

For conceptual knowledge representation to be useful,
we need to develop some practical guides for conceptual scaling.
From the user's standpoint,
there must be a {\em purpose\/} in mind
and an intended {\em use\/} for the information.
It would be good to write these down explicitly.
The information itself is usually concerned with {\em entities\/},
although entity tuples might be appropriate.
In traditionally crisp or hard conceptual knowledge representation,
in order to form a base for conceptual knowledge
we must ask {\em true-false questions\/} about the information.
We can compile these,
for purpose of interactive management,
into a database of natural language queries.
The bottom level of this query database forms coherent components
which we call {\em conceptual scales\/}.
This partitions the queries.
A conceptual scale is associated with a composite description function.
As listed in Table~\ref{scales},
conceptual scales can themselves be typed according to purpose or use, and mathematical structure \cite{ganter89}.
Mathematical types of scales represent intuitive ideas of design.

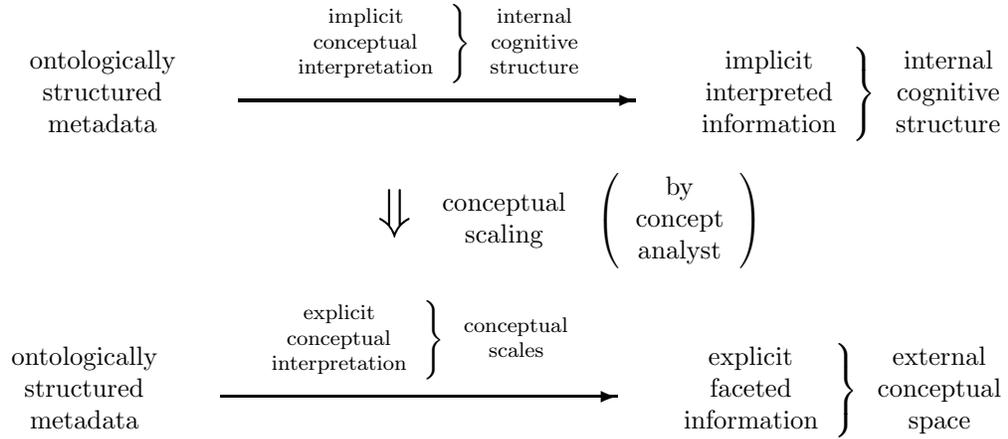
\begin{figure}
\begin{tabular}{c} \\ \mbox{ }
\begin{tabular}{rcl}
\begin{tabular}{c}
	ontologically \\
	structured \\
	metadata
\end{tabular}
&
\footnotesize
\begin{tabular}[b]{c}
$\left.
\mbox{\begin{tabular}{c}
	implicit \\
	conceptual \\
	interpretation
\end{tabular}}
\right\}
\mbox{\begin{tabular}{c}
	internal \\
	cognitive \\
	structure
\end{tabular}}$
\\
\begin{picture}(150,0)
	\thicklines
	\put(0,0){\vector(1,0){150}}
	\thinlines
\end{picture}
\end{tabular}
\normalsize
&
$\left.
\mbox{\begin{tabular}{c}
	implicit \\
	interpreted \\
	information
\end{tabular}}
\right\}
\mbox{\begin{tabular}{c}
	internal \\
	cognitive \\
	structure
\end{tabular}}$
\end{tabular}
\\
\hspace{45pt}
\begin{tabular}{c@{\hspace{2mm}}r@{\hspace{2mm}}l}
\\
\huge{$\Downarrow$}
&
\begin{tabular}{c}
	conceptual \\
	scaling 
\end{tabular}
&
$\left(
\mbox{\begin{tabular}{c}
	by \\
	concept \\
	analyst
\end{tabular}}
\right)$
\\ \\
\end{tabular}
\\
\begin{tabular}{rcl}
\begin{tabular}{c}
	ontologically \\
	structured \\
	metadata
\end{tabular}
&
\footnotesize
\begin{tabular}[b]{c}
$\left.
\mbox{\begin{tabular}{c}
	explicit \\
	conceptual \\
	interpretation
\end{tabular}}
\right\}
\mbox{\begin{tabular}{c}
	conceptual \\
	scales
\end{tabular}}$
\\
\begin{picture}(150,0)
	\thicklines
	\put(0,0){\vector(1,0){150}}
	\thinlines
\end{picture}
\end{tabular}
\normalsize
&
$\left.
\mbox{\begin{tabular}{c}
	explicit \\
	faceted \\
	information
\end{tabular}}
\right\}
\mbox{\begin{tabular}{c}
	external \\
	conceptual \\
	space
\end{tabular}}$
\end{tabular} \\ \mbox{ }
\end{tabular}
\caption{{\bf The process of conceptual scaling}}
\label{scaling}
\end{figure}
\begin{figure}
\begin{tabular}{ccccc}
\scriptsize{\begin{tabular}[t]{|c|c|} \hline
\multicolumn{2}{|c|}{\small\shortstack{{\bf description}\\{\bf function}}} \\ \hline\hline
	\small{\bf Person} & \small{\bf age} \\ \hline
	Adam	& 21 \\
	Betty	& 50 \\
	Chris	& 66 \\
	Dora	& 88 \\
	Eva	& 17 \\
	Fred	& ?  \\
	George	& 90 \\
	Harry	& 50 \\
	\hline
\end{tabular}}
&
\begin{tabular}[t]{c}
\\ \\ \\ \\ \\
\begin{picture}(30,0)(0,0)\thicklines\put(5,0){\vector(1,0){20}}\end{picture}
\end{tabular}
&
\begin{tabular}[t]{|c|} \hline
{\bf concrete conceptual scale}
\\ \hline\hline
\setlength{\unitlength}{1pt}
\begin{picture}(100,100)(15,10)
	\puttext{0}{0}{{\bf }}
	\putdisk{50}{100}{7}				
	\putdisk{20}{80}{7}				
	\puttext{25}{84}{{\rm working}}			
	\puttext{25}{75}{{\leq}65,{\geq}40}		
	\put(20,80){\line(3,2){30}}			
	\putdisk{80}{70}{7}				
	\puttext{85}{74}{{\rm retired}}			
	\puttext{85}{65}{{>}65,{<}80}			
	\put(80,70){\line(-1,1){30}}			
	\putdisk{20}{60}{7}				
	\puttext{25}{64}{{\rm young}}			
	\puttext{25}{55}{{<}40,{\geq}16}		
	\put(20,60){\line(0,1){20}}			
	\putdisk{80}{50}{7}				
	\puttext{85}{54}{{\rm old}}			
	\puttext{85}{45}{{\geq}80}			
	\put(80,50){\line(0,1){20}}			
	\putdisk{20}{40}{7}				
	\puttext{25}{44}{{\rm minor}}			
	\puttext{25}{35}{{<}18}				
	\put(20,40){\line(0,1){20}}			
	\putdisk{50}{20}{7}				
	\put(50,20){\line(-3,2){30}}			
	\put(50,20){\line(1,1){30}}			
\end{picture}
\\ \hline
\end{tabular}
&
\begin{tabular}[t]{c}
\\ \\ \\ \\ \\
\begin{picture}(30,0)(0,0)\thicklines\put(5,0){\vector(1,0){20}}\end{picture}
\end{tabular}
&
\begin{tabular}[t]{|c|} \hline
{\bf facet}
\\ \hline\hline
\setlength{\unitlength}{1pt}
\begin{picture}(100,100)(10,10)
	\puttext{0}{0}{{\bf }}
	\putdisk{50}{100}{7}				
	\puttext{55}{95}{{\rm Fred}}			
	\putdisk{20}{80}{7}				
	\puttext{25}{84}{{\rm working}}			
	\puttext{25}{75}{{\rm Betty}}			
	\puttext{25}{70}{{\rm Harry}}			
	\put(20,80){\line(3,2){30}}			
	\putdisk{80}{70}{7}				
	\puttext{85}{74}{{\rm retired}}			
	\puttext{85}{65}{{\rm Chris}}			
	\put(80,70){\line(-1,1){30}}			
	\putdisk{20}{60}{7}				
	\puttext{25}{64}{{\rm young}}			
	\puttext{25}{55}{{\rm Adam}}			
	\put(20,60){\line(0,1){20}}			
	\putdisk{80}{50}{7}				
	\puttext{85}{54}{{\rm old}}			
	\puttext{85}{45}{{\rm Dora}}			
	\puttext{85}{40}{{\rm George}}			
	\put(80,50){\line(0,1){20}}			
	\putdisk{20}{40}{7}				
	\puttext{25}{44}{{\rm minor}}			
	\puttext{25}{35}{{\rm Eva}}			
	\put(20,40){\line(0,1){20}}			
	\putdisk{50}{20}{7}				
	\put(50,20){\line(-3,2){30}}			
	\put(50,20){\line(1,1){30}}			
\end{picture}
\\ \hline
\end{tabular}
\end{tabular}
\caption{{\bf Conceptual scaling process: {\tt Age}}}
\label{interpretation}
\end{figure}
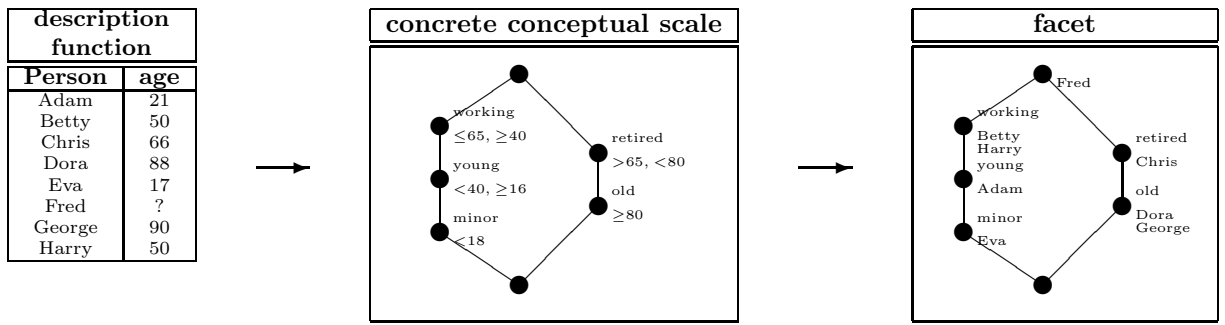

The process of conceptual scaling as depicted in Figure~\ref{scaling}
consists of the interpretation of ontologically structured metadata.
From a technical standpoint as depicted in Figure~\ref{interpretation},
conceptual scaling is the conversion of a composite description function to a facet.
But more importantly from both philosophical and practical standpoints,
the development of conceptual scales is an act of interpretation
which defines views of the information along a variety of information dimensions called facets.
These facets form a spectrum of interpretation/classification,
from the very particular and often ad hoc,
through the more pragmatic and utilitarian,
to the very general and scientific.


There are three constituents in the development of facets by means of conceptual scales:
the abstract conceptual scale, 
the concrete conceptual scale, 
and the process of conceptual scaling.

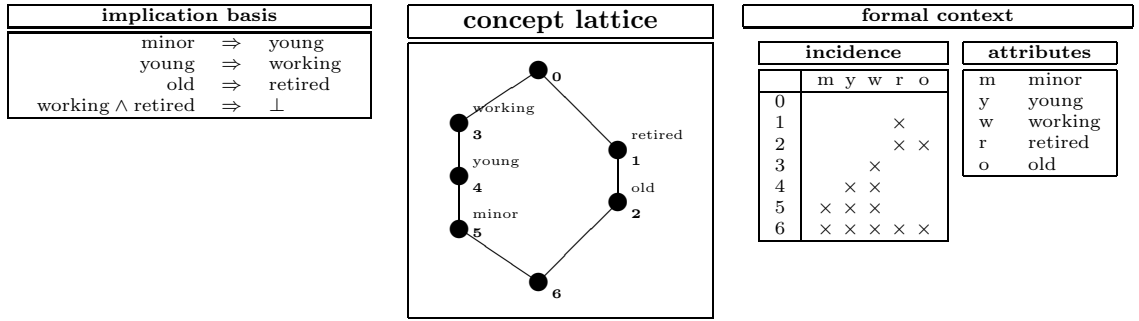
\begin{figure}
\begin{center}
\begin{tabular}{c@{\hspace{4mm}}c@{\hspace{4mm}}c}
{\scriptsize
\begin{tabular}[t]{|c|} \hline
{\bf implication basis}
\\ \hline\hline
$\begin{array}{rcl}
	{\rm minor}   & \Rightarrow & {\rm young}     \\
	{\rm young}   & \Rightarrow & {\rm working}   \\
	{\rm old} & \Rightarrow & {\rm retired}   \\
	{\rm working} \wedge {\rm retired} & \Rightarrow & \bot
\end{array}$
\\ \hline
\end{tabular}
\normalsize}
&
\begin{tabular}[t]{|c|} \hline
{\bf concept lattice}
\\ \hline\hline
\setlength{\unitlength}{1pt}
\begin{picture}(100,100)(10,10)
	\puttext{0}{0}{{\bf }}
	\putdisk{50}{100}{7}				
	\puttext{55}{96}{{\bf 0}}			
	\putdisk{20}{80}{7}					
	\puttext{25}{84}{{\rm working}}		
	\puttext{25}{74}{{\bf 3}}			
	\put(20,80){\line(3,2){30}}			
	\putdisk{80}{70}{7}					
	\puttext{85}{74}{{\rm retired}}		
	\puttext{85}{64}{{\bf 1}}			
	\put(80,70){\line(-1,1){30}}		
	\putdisk{20}{60}{7}					
	\puttext{25}{64}{{\rm young}}		
	\puttext{25}{54}{{\bf 4}}			
	\put(20,60){\line(0,1){20}}			
	\putdisk{80}{50}{7}					
	\puttext{85}{54}{{\rm old}}		
	\puttext{85}{44}{{\bf 2}}			
	\put(80,50){\line(0,1){20}}			
	\putdisk{20}{40}{7}					
	\puttext{25}{44}{{\rm minor}}		
	\puttext{25}{37}{{\bf 5}}			
	\put(20,40){\line(0,1){20}}			
	\putdisk{50}{20}{7}					
	\puttext{55}{14}{{\bf 6}}			
	\put(50,20){\line(-3,2){30}}		
	\put(50,20){\line(1,1){30}}			
\end{picture}
\\ \hline
\end{tabular}
&
{\scriptsize
\begin{tabular}[t]{c@{\hspace{2mm}}c} \hline
	\multicolumn{2}{|c|}{{\bf formal context}} \\ \hline
	\begin{tabular}[t]{|l|c@{\hspace{3pt}}c@{\hspace{3pt}}c@{\hspace{3pt}}c@{\hspace{3pt}}c|} \hline
		\multicolumn{6}{|c|}{{\bf incidence}} \\ \hline\hline
		   & m & y & w & r & o \\ \hline
		0 &&&&& \\
		1 &&&&$\times$& \\
		2 &&&&$\times$&$\times$ \\
		3 &&&$\times$&& \\
		4 &&$\times$&$\times$&& \\
		5 &$\times$&$\times$&$\times$&& \\
		6 &$\times$&$\times$&$\times$&$\times$&$\times$ \\ \hline
	\end{tabular}
	&
	\begin{tabular}[t]{|ll|} \hline
		\multicolumn{2}{|c|}{{\bf attributes}} \\ \hline\hline
		m & minor    \\
		y & young    \\
		w & working  \\
		r & retired  \\
		o & old  \\ \hline
	\end{tabular}
\end{tabular}
\normalsize}
\end{tabular}
\end{center}
\caption{{\bf Abstract conceptual scale: {\tt Age} (3 forms)}}
\label{Age:scheme}
\end{figure}

\begin{itemize}
\item {\bf [Abstract conceptual scale]} 
	The first constituent is 
	the development of a conceptual scale associated with a composite description function.
	This involves the linguistic analysis of this information dimension.
	\begin{enumerate}
	\item The creation and choice of scale attributes or linguistic values (terms)
		--- words or phrases meaningful for this particular information dimension.
		It is very important to observe that terms form a spectrum,
		from terms used for very individualistic and ad hoc interpretation/classification
		to 
		terms with a common and accepted meaning in science and society.
		\begin{center}
		\begin{tabular}{c@{\hspace{15mm}}c@{\hspace{15mm}}c}
			\begin{tabular}[b]{c} {\tt young}, \\ {\tt old} \end{tabular}
				& \begin{tabular}[b]{c} {\tt working} \end{tabular}
					& \begin{tabular}[b]{c} {\tt minor}, \\ {\tt retired} \end{tabular}
			\\
			\multicolumn{3}{c}{\begin{picture}(220,0)(0,-2)\thicklines
			\put(-20,0){\vector(1,0){260}}
			\put(-20,0){\vector(-1,0){0}}
			\end{picture}}
			\\
			{\it individual\/} & {\it pragmatic\/} & {\it standard\/}
		\end{tabular}
		\end{center}

	\item The analysis of implicational structure between terms.
		As an aid in the explication of conceptual scaling we will use our adaptation of 
		an example of people's age developed in an unpublished report by Karl Erich Wolff.
		The abstract {\tt Age} conceptual scale is represented (equivalently) in Figure~\ref{Age:scheme} 
		as either a basis of implications, a lattice, or a one-valued formal context.
		The Person ontology in Table~\ref{CKML}
		specifies the abstract {\tt Age} conceptual scale in CKML.
		This particular abstract {\tt Age} conceptual scale has the form of a biordinal scale.
		The total set of implications can be generated from the basis of implications 
		by use of the following inference rules.
		\begin{scriptsizeflushleft}
		\begin{tabular}{c@{\hspace{3mm}}c@{\hspace{3mm}}c} \\
		\begin{tabular}{lc}
		{\bf transitive:} &
		\begin{tabular}{c}
			$X \Rightarrow Y$, $Y \Rightarrow Z$
			\\ \hline
			$X \Rightarrow Z$
		\end{tabular}
		\end{tabular}
		&
		\begin{tabular}{lc}
		{\bf projective:} &
		\begin{tabular}{c}
			$X \supseteq Y$
			\\ \hline
			$X \Rightarrow Y$
		\end{tabular}
		\end{tabular}
		&
		\begin{tabular}{lc}
		{\bf additive:} &
		\begin{tabular}{c}
			$X \Rightarrow Y$, $X \Rightarrow Z$
			\\ \hline
			$X \Rightarrow (Y \cup Z)$
		\end{tabular}
		\end{tabular} \\ \\ \\
		\end{tabular}
		\end{scriptsizeflushleft}
\end{enumerate}
\item {\bf [Concrete conceptual scale]} 
	The second constituent is 
	the development of a concrete conceptual scale over the natural numbers primitive data type
	$D = \{0,1,2,\cdots,\} = \aleph$.
	This involves the binding of the abstract conceptual scale with logical queries.
	\begin{enumerate}
	\item Assignment of logical query formula to the terms of the abstract {\tt Age} conceptual scale
		as in Table~\ref{assignments}.
		The logical query formulas must satisfy the constraints listed in Table~\ref{constraints}
		specified by the implicational basis of the abstract {\tt Age} conceptual scale.

\begin{table}
\begin{center}
		\footnotesize
		$\left.
		\mbox{\begin{tabular}{rclcl}
			{\tt minor}   & $\longmapsto$ & $\phi_{\rm minor}$   & = & $(x \leq 18)$? \\
			{\tt young}   & $\longmapsto$ & $\phi_{\rm young}$   & = & $(x < 40)?$    \\
			{\tt working} & $\longmapsto$ & $\phi_{\rm working}$ & = & $(x \leq 65)?$ \\
			{\tt retired} & $\longmapsto$ & $\phi_{\rm retired}$ & = & $(x > 65)?$    \\
			{\tt old} & $\longmapsto$ & $\phi_{\rm old}$ & = & $(x \geq 80)?$
		\end{tabular}}
		\right\}
		\mbox{assignment}$
		\normalsize
\caption{{\bf Conceptual scale assignment}}
\label{assignments}
\end{center}
\end{table}

\begin{table}
\begin{center}
		\footnotesize
		$\left.
		\begin{array}{rcl}
			\phi_{\rm minor}   & \subseteq & \phi_{\rm young}   \\
			\phi_{\rm young}   & \subseteq & \phi_{\rm working} \\
			\phi_{\rm old} & \subseteq & \phi_{\rm retired} \\
			\phi_{\rm working} \cap \phi_{\rm retired} & = & \emptyset
		\end{array}
		\right\}
		\mbox{\begin{tabular}{l}implicational\\constraint\\satisfaction\end{tabular}}$
		\normalsize
\caption{{\bf Conceptual scale constraints}}
\label{constraints}
\end{center}
\end{table}

	\item Calculation of conceptual contingents (distinguishing characteristics).
		Recall the structure of the usual dictionary definition
		in terms of superordinate concept and distinguishing characteristics.
		For example,
		the definiton of the category of trees given as follows.
		\begin{quotation}
		\begin{description}
			\item[tree] {\em noun}: a woody perennial plant 
				having a single usually elongated main stem 
				generally with few or no branches on its lower part
		\end{description}
		\end{quotation}
		Here the category of trees has the category of plants as its immediate superordinate,
		and has the distinguishing characteristics:
		woody, perennial, one-branching-stem.
		The concrete {\tt Age} conceptual scale is represented as a lattice 
		in the center of Figure~\ref{interpretation}
		by calculating concept contingents $\gamma_n$ for each concept $n$
		via the formula
		\begin{center}
			$\gamma_n \;\;\define\;\; \phi_n \wedge ( \bigwedge_i \neg \phi_i )$,
		\end{center}
		where $i$ ranges over all children nodes of $n$.
		Together with the Person ontology, the Person collection in Table~\ref{CKML} 
		specifies the concrete {\tt Age} conceptual scale in CKML
		by the assignment of queries to terms of the {\tt Age} scale in the Person ontology.
	\end{enumerate}
\item {\bf [Conceptual scaling process]} 
	The third constituent is
	the evaluation of logical queries with respect to ontologically structured metadata.
	Facet interpretation of metadata involves contingent query evaluation.
	Consider the {\tt People} category,
	described on the left side in Figure~\ref{interpretation},
	which might be part of a questionaire.
	Missing values, so-called database nulls,
	are represented by the question mark ``{?}''.
	The {\tt Age} dimension of the {\tt People} data
	is interpreted in terms of the {\tt Age} conceptual scale
	by evaluating the concept contingent logical query.
	These evaluations interpret the information as a single {\tt Age} facet,
	which can be visualized as the concept lattice 
	on the right side in Figure~\ref{interpretation}.
\end{itemize}

\begin{table}
\begin{tabular}[t]{cc}

\begin{tabular}[t]{|c|} \hline \\ \begin{minipage}{2.25in}\tiny\begin{verbatim}
   <ONTOLOGY NAME="Person" VERSION="1.0">
      ...  
      <CATEGORY NAME="Person"/>   
      ...  
      <FNSCHEMA NAME="age" 
         ARGTYPE="Person" 
         IMAGETYPE="Integer"/>  
      ...  
      <SCALE CATEGORY="Person" NAME="Age">    
         <TERM NAME="Young"/>
         <TERM NAME="Old"/>
         <TERM NAME="Working"/>
         <TERM NAME="Minor"/>
         <TERM NAME="Retired"/>
         <IMPLICATION>
            <IF><TERM NAME="Minor/></IF>
            <THEN><TERM NAME="Young"/></THEN>
         </IMPLICATION>
         <IMPLICATION>
            <IF><TERM NAME="Young/></IF>
            <THEN><TERM NAME="Working"/></THEN>
         </IMPLICATION>
         <IMPLICATION>
            <IF><TERM NAME="Old/></IF>
            <THEN><TERM NAME="Retired"/></THEN>
         </IMPLICATION>
         <IMPLICATION>
            <IF><TERM NAME="Working"/>
                <TERM NAME="Retired"/></IF>
         </IMPLICATION>
      </SCALE>
      ...  
   </ONTOLOGY>
\end{verbatim}\normalsize\end{minipage} \\ \\ \hline \end{tabular}
&
\begin{tabular}[t]{|c|} \hline \\ \begin{minipage}{2.25in}\tiny\begin{verbatim}
   <COLLECTION KIND="attribute" SCOPE="People">     
      <USES ONTOLOGY="Person" VERSION="1.0"/> 
      ...  
      <ATTRIBUTE SCALE="Age" KEY="Minor">
         <QUERY VARIABLE="person" CATEGORY="People"/>
            <FN2REL NAME="age" ORDER="less-equal">
               <ARGUMENT VALUE="person"/>
               <ARGUMENT VALUE="18"/>
            </FN2REL>
         </QUERY>
      </ATTRIBUTE>
      ...  
   </COLLECTION>
\end{verbatim}\normalsize\end{minipage} \\ \\ \hline \end{tabular}
\end{tabular}
\caption{{\bf Ontology and attribute collection in CKML}}
\label{CKML}
\end{table}

\section{Valuated Enrichment}

Indiscernibility,
a central concept in rough set theory,
is traditionally treated as a hard relationship
--- either two objects are indiscernible 
or they are not.
In order to define and develop a soft theory of rough sets,
it would seem quite appropriate,
if not necessary,
to define and develop a soft or graded version of indiscernibility.
We review this approach\cite{kent94} by using ideas from the theory of valuated categories\cite{lawvere73}.

An {\em approximation space\/} \cite{pawlak82} is traditionally defined as
a pair ${\cal G} = \pair{G}{{E}}$
consisting of
a set of objects or entities $G$
and 
an equivalence relation $E \subseteq \product{G}{G}$
called indiscernibility.
Two objects $g_1,g_2 \memberof G$ are {\em indiscernible\/} 
when $g_1{E}g_2$;
that is,
when $E(g_1,g_2) = {\tt true}$.
Equivalently,
an approximation space (function version) is a triple $\triple{G}{\phi}{D}$,
where $G$ is a set of objects,
$D$ is a set (hard, crisp and unenriched!) of values,
and $G \stackrel{\phi}{\rightarrow} D$ is a (not necessarily surjective) function 
called a {\em description function\/}.
The description function $\phi$ represents a certain amount of knowledge 
about the objects in $G$.
Two objects $g_1,g_2 \memberof G$ are {\em indiscernible\/}
when 
the procedure $\phi$ cannot distinguish between them,
$\phi(g_1) = \phi(g_2)$;
that is,
when ${\rm Eq}_D(\phi(g_1),\phi(g_2)) = {\tt true}$.
To enriched rough set notions,
we allow grades of indiscernibility
by assuming that $D$ has {\bf V}-enriched structure on it,
where {\bf V} = $\quintuple{V}{\preceq}{\tensor}{\implication}{e}$
is a closed preorder;
that is,
we assume that $D$ is an approximation {\bf V}-space.

\label{closed_preorder}

A {\em closed preorder}
\cite{lawvere73}
${\bf V} = \quintuple{V}{\preceq}{\tensor}{\implication}{e}$
consist of the following data and axioms.
$\quadruple{V}{\preceq}{\tensor}{e}$ is a symmetric monoidal preorder,
with $\pair{V}{\preceq}$ a preorder and $\triple{V}{\tensor}{e}$ a commutative monoid,
where the binary operation $\tensor \morph \product{V}{V} \rightarrow V$, 
called {\bf V}-{\em composition\/}, is monotonic
(if both $u \preceq u'$ and $v \preceq v'$ then $u \tensor v \preceq u' \tensor v'$),
and symmetric or commutative
($a \tensor b = b \tensor a$ for all elements $a,b \in V$),
and satisfies the closure axiom:
the monotonic {\bf V}-composition function
$(\:) \tensor b \morph V \rightarrow V$
has a specified right adjoint
$b \implication (\:) \morph V \rightarrow V$
for each element $b \in B$
called {\bf V}-{\em implication\/},
hence satisfying the equivalence
$a \tensor b \preceq c$ iff $a \preceq b \implication c$
for any triple of elements $a,b,c \in V$.
We list some important closed preorders 
which can be used in soft concept analysis for the interpretation in linguistic variables.
\begin{scriptsizedescription}
	\item[booleans]
		$\quintuple{2=\{0,1\}}{\leq}{\wedge}{\rightarrow}{1}$,
		where 0 is {\tt false}, 1 is {\tt true}, $\leq$ is the usual order on truth-values,
		$\wedge$ is the truth-table for {\tt and}, and $\rightarrow$ is the truth-table for {\tt implies}.
		This defines the hard context of traditional set theory and logic.
	\item[fuzzy truth-values]
		$\quintuple{[0,1]}{\leq}{\wedge}{\rightarrow}{1}$
		where 0 is {\tt false}, 1 is {\tt true}, 
		$0 \leq r \leq 1$ is some grade of truth-value between {\tt false} and {\tt true},
		$\leq$ is the usual order on fuzzy truth-values in the interval,
		$\wedge$ is the minimum operation representing the interval truth-table 
		for the fuzzy {\tt and},
		and $\rightarrow$ is the operation
		(defined by $r \rightarrow s = 1$ if $r \leq s$, or $s$ otherwise)
		representing the interval truth-table for the fuzzy {\tt implies}.
		This defines a soft context for fuzzy set theory and logic.
	\item[reals]
		$\Re$ = $\quintuple{\Re=[0,\infty]}{\geq}{+}{\minus}{0}$,
		$\geq$ is the usual downward ordering on the nonegative real numbers $\Re$
		(regarded as quantitative truth-values),
		$+$ is sum, 
		and $\minus$ is the operation
		(defined by $s \minus r = 0$ if $r \geq s$, or $s - r$ otherwise)
		representing the truth-table for the metrical {\tt difference}.
		This defines the soft context of metric spaces.
\end{scriptsizedescription}

In soft concept analysis the operation of {\bf V}-implication is used for at least three different purposes:
(1) the enriched lower approximation operator uses {\bf V}-implication;
(2) {\bf V}-implication is sometimes used in the queries which define enriched conceptual scales;
and
(3) the operation of derivation, which defines the notion of an enriched formal concept, 
is a special case of enriched relational residuation,
which itself is defined using {\bf V}-implication.
The values in the closed preorder $V$ are regarded as being a set of generalized truth values.


While enriched approximation spaces are the appropriate abstraction of indiscernibility
and our main concern in this paper,
it seems that these approximation spaces are best defined in terms of 
an asymmetric generalization called simply an enriched space.
A pair ${\cal X} = \pair{X}{\mu}$ consisting of 
a set $X$ and a function $\mu \morph \product{X}{X} \rightarrow {\bf V}$ 
is called a {\em {\bf V}-enriched space\/} or {\em {\bf V}-space}
when it satisfies 
the reflexivity (zero law) 
$e \preceq \mu(x,x)$ for all $x \memberof X$,
and
the transitivity (triangle axiom) 
$\mu(x_1,x_2) \tensor \mu(x_2,x_3) \preceq \mu(x_1,x_3)$ for all $x_1,x_2,x_3 \memberof X$.
The function $\mu$, called a {\em metric\/},
represents a distance or measure of agreement between the elements of $X$.
We can interpret $\mu$ to be 
either 
an enriched preordering,
a generalized distance function,
a similarity measure,
or 
a gradation.
When {\bf V} = 2, the crisp boolean case,
a {\bf V}-space ${\cal X}$ is precisely a preorder 
${\cal X} = \pair{X}{\preceq}$
with order characteristic function
${\preceq} \morph \product{X}{X} \rightarrow 2$.
When {\bf V} = $\Re$, the metric topology case,
a {\bf V}-space ${\cal X}$ is (generalize) metric space 
${\cal X} = \pair{X}{\delta}$
with distance function
$\delta \morph \product{X}{X} \rightarrow \Re$.
When {\bf V} = $[0,1]$, the fuzzy case,
a {\bf V}-space ${\cal X}$ is a fuzzy space 
${\cal X} = \pair{X}{\mu}$
with similarity measure
$\mu \morph \product{X}{X} \rightarrow [0,1]$.

Any {\bf V}-space ${\cal X} = \pair{X}{\mu}$ has 
a {\em dual\/} or {\em opposite\/} {\bf V}-space 
${\cal X}^{\rm op} = \pair{X}{\mu^{\rm op}}$,
where $\mu^{\rm op}(x_1,x_2) = \mu(x_2,x_1)$ is the dual or opposite metric.
In general our metrics are asymmetrical:
$\mu(x_1,x_2) \neq \mu(x_2,x_1)$.
A {\em {\bf V}-enriched approximation space\/} 
or {\em approximation {\bf V}-space\/}
is defined to be a symmetrical {\bf V}-space.
Here the metric $\mu$,
called an {\em indiscernibility measure\/},
is a {\bf V}-enriched equivalence relation on $X$ 
satisfying reflexivity, transitivity and
symmetry 
$\mu(x_2,x_1) = \mu(x_1,x_2)$ for all $x_1,x_2 \memberof X$.
Any {\bf V}-space ${\cal X} = \pair{X}{\mu}$ can be symmetrized 
and made into an approximation space,
by defining the metric
$\mu^{\rm sym}(x_1,x_2) = \mu(x_1,x_2) \tensor \mu^{\rm op}(x_1,x_2)$.

A {\bf V}-{\em map} 
$f \morph {\cal X} \rightarrow {\cal Y}$
between two {\bf V}-spaces 
${\cal X} = \pair{X}{\mu}$ and ${\cal Y} = \pair{Y}{\nu}$
is a function $f \morph X \rightarrow Y$ 
that preserves measure
by satisfying the condition
$\mu(x_1,x_2) \preceq \nu(f(x_1),f(x_2))$ for all $x_1,x_2 \in X$.
When {\bf V} = 2, the crisp boolean case,
a {\bf V}-map $f \morph {\cal X} \rightarrow {\cal Y}$ 
is precisely a monotonic function.
When {\bf V} = $\Re$, the metric topology case,
a {\bf V}-map $f \morph {\cal X} \rightarrow {\cal Y}$ 
is precisely a contraction.
When {\bf V} = $[0,1]$, the fuzzy case,
a {\bf V}-map $f \morph {\cal X} \rightarrow {\cal Y}$ 
is a fuzzy measure preserving function.


Each element $x \memberof X$ of a {\bf V}-space ${\cal X} = \pair{X}{\mu}$
can be represented as the {\bf V}-predicate ${\rm y}(x) = \mu(x,-)$ over ${\cal X}$
where ${\rm y}(x)(x') = \mu(x,x')$ for each element $x' \memberof X$.
The function ${\rm y}_X \morph X \rightarrow V^X$,
which is called the {\em Yoneda embedding},
is a {\bf V}-isometry 
${\rm y}_{\cal X} \morph {\cal X}^{\rm op} \rightarrow {\bf V}^{\cal X}$.
Composition of (the opposite of) a {\bf V}-map
$f \morph {\cal X} \rightarrow {\cal Y}$
on the right with the Yoneda embedding
${\rm y}_{\cal Y} \morph {\cal Y}^{\rm op} \rightarrow {\bf V}^{\cal Y}$,
resulting in the {\bf V}-map
$f_\ast \morph {\cal X}^{\rm op} \rightarrow {\bf V}^{\cal Y}$,
allows us to generalize the concept of a {\bf V}-map.
Such a generalized {\bf V}-map,
equivalent to a {\bf V}-map
$\product{{\cal X}^{\rm op}}{{\cal Y}} \stackrel{\tau}{\rightarrow} {\bf V}$,
may be regarded to be 
a {\em {\bf V}-enriched relation\/} or {\em {\bf V}-relation\/} 
from ${\cal X}$ to ${\cal Y}$.
It is denoted by ${\cal X} \stackrel{\tau}{\rightharpoondown} {\cal Y}$,
with $\tau(x,y)$ an element of {\bf V} interpreted as the
``truth-value of the $\tau$-relatedness of $x$ to $y$'' \cite{lawvere73}.

A pair of {\bf V}-relations 
${\cal X} \stackrel{\sigma}{\rightharpoondown} {\cal Y}$ 
and 
${\cal Y} \stackrel{\tau}{\rightharpoondown} {\cal Z}$
can be composed,
yielding the {\bf V}-relation 
${\cal X} \stackrel{\sigma\circ\tau}{\rightharpoondown} {\cal Z}$
called {\em composition\/},
and defined to be the supremum (iterated disjunction)
$(\sigma \circ \tau)(x,z) = \bigvee_{y \in {\cal Y}} \left( \sigma(x,y) \tensor \rho(y,z) \right)$.
An enriched relation ${\cal X} \stackrel{\tau}{\rightharpoondown} {\cal Y}$
is closed with respect to the metrics on both left and right:
$\mu \circ \tau \preceq \tau$ and $\tau \circ \nu \preceq \tau$.
Relational composition has a right adjoint called residuation.
The {\em residuation\/} of a pair of {\bf V}-relations 
${\cal X} \stackrel{\sigma}{\rightharpoondown} {\cal Y}$ 
and 
${\cal X} \stackrel{\rho}{\rightharpoondown} {\cal Z}$,
denoted by the {\bf V}-relation 
${\cal Y} \stackrel{\sigma\tensorimplysource\rho}{\rightharpoondown} {\cal Z}$,
is defined to be the infimum (iterated conjunction)
$(\sigma \tensorimplysource \rho)(y,z) = \bigwedge_{x \in {\cal X}} \left( \sigma(x,y) \implication \rho(x,z) \right)$.

As mentioned above,
every {\bf V}-map 
${\cal X} \stackrel{f}{\rightarrow} {\cal Y}$
determines a {\bf V}-relation 
${\cal X} \stackrel{f_\ast}{\rightharpoondown} {\cal Y}$
defined by $f_\ast = f^{\rm op} \cdot {\rm y}_{\cal Y}$,
or on elements by $f_\ast(x,y) = \nu(f(x),y)$.
In particular,
the Yoneda embedding becomes the relation ${\cal X} \stackrel{\mu}{\rightharpoondown} {\cal X}$.
Dually every {\bf V}-map 
${\cal X} \stackrel{f}{\rightarrow} {\cal Y}$ 
also determines a {\bf V}-relation 
${\cal Y} \stackrel{f^\ast}{\rightharpoondown} {\cal X}$
in the opposite direction
defined by $f^\ast = {\rm y}_{\cal Y} \cdot {\bf V}^f$,
or on elements by $f^\ast(y,x) = \nu(y,f(x))$.

The {\em power {\bf V}-space\/} ${\bf V}^{{\cal X}}$
of all {\bf V}-valued {\bf V}-maps on ${\cal X}$
is an {\bf V}-space with metric
$\phi \implication \psi = \bigwedge_{x \in X} \left( \phi(x) \implication \psi(x) \right)$.
We interpret an element of ${\bf V}^{\cal X}$,
a {\bf V}-map $\phi \morph {\cal X} \longrightarrow {\bf V}$,
to be a {\bf V}-enriched subset,
which satisfies the internal pointwise metric constraint $\mu$:
$\mu(x_1,x_2) \preceq \phi(x_1) \implication \phi(x_2)$ for all $x_1,x_2 \in X$;
or equivalently,
by the $\tensor$ - $\implication$ adjointness,
the constraint
$\phi(x_1) \tensor \mu(x_1,x_2) \preceq \phi(x_2)$ for all $x_1,x_2 \memberof X$.
Such a characteristic function $\phi \morph {\cal X} \rightarrow {\bf V}$,
which is constrained by the metric on ${\cal X}$,
is called a {\em {\bf V}-predicate\/} or {\em enriched predicate\/} in ${\cal X}$.
Using terminology from rough set theory,
it can also be called a {\em {\bf V}-definable subset\/} in ${\cal X}$.

When {\bf V} = 2, the crisp boolean case, for an approximation space ${\cal X} = \pair{X}{E}$,
a {\bf V}-predicate $\phi \morph {\cal X} \rightarrow 2$ 
satisfying the constraint
$\phi(x_1) \wedge E(x_1,x_2) \leq \phi(x_2)$ for all $x_1,x_2 \memberof X$
is precisely a definable subset in $X$.
When {\bf V} = $\Re$, the metric topology case, for an {\bf V}-space ${\cal X} = \pair{X}{\delta}$,
a {\bf V}-predicate $\phi \morph {\cal X} \rightarrow \Re$ 
satisfying the constraint
$\phi(x_1) + \delta(x_1,x_2) \geq \phi(x_2)$ for all $x_1,x_2 \memberof X$
is called a {\em closed subset\/} of ${\cal X}$
--- closed w.r.t.\ the distance function $\delta$.
When {\bf V} = $[0,1]$, the fuzzy case,
a {\bf V}-predicate $\phi \morph {\cal X} \rightarrow [0,1]$
satisfying the constraint
$\phi(x_1) \wedge \mu(x_1,x_2) \leq \phi(x_2)$ for all $x_1,x_2 \memberof X$,
or equivalently,
the constraint ``equal below points of similarity''
$\mu(x_1,x_2) \leq \phi(x_1), \phi(x_2)$
or
$\phi(x_1) = \phi(x_2) \leq \mu(x_1,x_2)$.


\section{Enriched Conceptual Scales}

We describe enriched conceptual scales ($\equiv$ enriched linguistic variables)
in terms of a use-case scenario.
We start with a collection of objects ${\cal G} = \pair{G}{\gamma}$.
We assume that some observations have been made
or some experimental measurements have been done,
resulting in the production of some ``raw'' data ${\cal D} = \pair{D}{\delta}$.
This data is associated with the objects by a map called a {\em description function\/}
${\cal G} \stackrel{\phi}{\rightarrow} {\cal D}$.
Both objects and data can be enriched as approximation spaces
for benefit of flexibility by using soft structures.
We will use enriched conceptual scales in order 
(1) to interpret this data 
and 
(2) to provide a facet of it which is meaningful to the user.
The creation of enriched conceptual scales is an act of interpretation.

Mathematically,
the notion of an enriched attribute ($\equiv$ linguistic value) is represented here 
by the notion of an enriched predicate.
An {\em enriched attribute\/} over data domain ${\cal D} = \pair{D}{\delta}$
is an enriched predicate in ${\bf V}^{\cal D}$.
An {\em enriched conceptual scale\/} ($\equiv$ {\em enriched linguistic variable\/})
\cite{zadeh75,ganter89,kent94}
over data domain ${\cal D} = \pair{D}{\delta}$
is a collection
$\sigma = \{ \sigma_m \in {\bf V}^{\cal D} \mid m \in M \}$
of enriched attributes over ${\cal D}$,
indexed by a collection of attribute symbols or terms $M$.
In the crisp case, {\bf V} = $2$,
the assignments are part of a concrete conceptual scale (see Table~\ref{assignments}).
Using functional notation
we can write this as the {\bf V}-map
$\sigma \morph {\cal M} \rightarrow {\bf V}^{\cal D}$,
where we have enriched the attributes to an {\bf V}-space ${\cal M} = \pair{M}{\mu}$.
In the crisp boolean case, {\bf V} = $2$,
the metric $\mu$ represents the implication basis order in an abstract conceptual scale (see Figure~\ref{Age:scheme}).
An enriched conceptual scale can be represented as the relation
${\cal M} \stackrel{\sigma}{\rightharpoondown} {\cal D}$
where $\sigma(m,d) \define \sigma(m)(d)$.
In the crisp boolean case, {\bf V} = $2$,
closure of this enriched relation with respect to the term metric 
$\mu \circ \sigma \preceq \sigma$ 
represents constraint satisfaction in the concrete conceptual scale (see Table~\ref{constraints}).
The four parts of a enriched conceptual scale can be interpreted as follows.
\begin{scriptsizeenumerate}
	\item ${\cal D}$ gives its data scope or range,
	\item ${\bf V}$ represents our interpretation style,
	\item ${\cal M}$ gives attributes of the enriched conceptual scale
	\item $\sigma$ assigns enriched predicates to terms.
\end{scriptsizeenumerate}
These are listed in order of volatility
--- of these four,
${\cal D}$ varies slowest (it is given to us),
whereas
$\sigma$ is most volatile.
The standard way to combine two enriched conceptual scales is used in the apposition of formal contexts \cite{ganter89}.
Given two enriched conceptual scales (with no apparent relationships)
${\cal M}_0 \stackrel{\sigma_0}{\rightharpoondown} {\cal D}_0$
and
${\cal M}_1 \stackrel{\sigma_1}{\rightharpoondown} {\cal D}_1$,
the {\em apposition\/} enriched conceptual scale
${\cal M}_0 \oplus {\cal M}_1 \stackrel{{\sigma_0}|{\sigma_1}}{\rightharpoondown} {\cal D}_0 \tensor {\cal D}_1$
from the unconstrained sum space of terms to the tensor product space of data,
is defined by
${\sigma_0}|{\sigma_1}\;(m_0,(d_0,d_1)) \define  \sigma_0(m_0,d_0)$
and
${\sigma_0}|{\sigma_1}\;(m_1,(d_0,d_1)) \define  \sigma_1(m_1,d_1)$. 

Using the {\tt Age} example discussed above for conceptual scaling,
we can provide a crisp interpretation of people's age description using the boolean closed poset
\begin{scriptsizeflushleft}
$\begin{array}{r@{\hspace{2mm}=\hspace{2mm}}l}
{\cal G} & {\tt Person}	\\
\phi     & \mbox{age description function}	\\
{\cal D} & {\bf \aleph} = \{0,1,2, \ldots \}	\\
{\bf V}	 & {\bf 2} = \mbox{crisp boolean closed poset}	\\
{\cal M} & \{\mbox{``minor''},\mbox{``young''},\mbox{``working''},\mbox{``retired''},\mbox{``old''}\}	\\
{\cal M} \stackrel{\sigma}{\rightharpoondown} {\cal D} &
\left[ \begin{array}{r@{\hspace{2mm}=\hspace{2mm}}l}
\sigma(\mbox{``young''})(d) &
	\left\{ \begin{array}{cl}
	1,				&	\;\;\mbox{if}\;\; 0  \leq d < 40	\\
	0,				&	\;\;\mbox{if}\;\; 40 \leq d
	\end{array} \right. \\
\multicolumn{1}{r}{\mbox{etc.}}
\end{array} \right]
\end{array}$
\end{scriptsizeflushleft}
or we can provide a fuzzy interpretation of people's age description using the fuzzy truth-values closed poset
and fuzzy predicates assigned to terms,
such as
\begin{scriptsizeflushleft}
$\left[ \begin{array}{r@{\hspace{2mm}=\hspace{2mm}}l}
\sigma(\mbox{``young''})(d)&
	\left\{ \begin{array}{cl}
	1,				&	\;\;\mbox{if}\;\; 0  \leq d \leq 20	\\
	-\frac{1}{20}d+2,		&	\;\;\mbox{if}\;\; 20 \leq d \leq 40	\\
	0,				&	\;\;\mbox{if}\;\; 40 \leq d
	\end{array} \right. \\
\multicolumn{1}{r}{\mbox{etc.}}
\end{array} \right]$
\end{scriptsizeflushleft}

We use an enriched conceptual scale to interpret the meaning of the metadata 
--- the description function $\phi$.
This interpretation,
called {\em simple enriched conceptual scaling\/},
applies the conceptual scale $\sigma$ by composing it with the description function metadata $\phi$,
resulting in the facet $\iota \define \phi_\ast \circ \sigma^{\rm op}$.
This takes the form of a {\bf V}-relation 
${\cal G} \stackrel{\iota}{\rightharpoondown} {\cal M}$
called an enriched formal context.
In terms of elements this definition is
$\iota(g,m) = \tilde{\sigma}(\phi(g))(m) = \sigma(m,\phi(g)) = \sigma^{op}(\phi(g),m)$.
It is important to observe that
simple enriched conceptual scaling is synonymous with the notion of {\em granulation\/} in fuzzy set theory.

Composite description functions within the metadata
allow for the definition of richer and more complex attributes in composite conceptual scales.
Such {\em composite enriched conceptual scaling\/} 
provides for automatic translation of natural language specifications\cite{woods91} of conceptual scales.
For example,
consider the composite description function
consisting of
a binary ``membership'' relation between the categories ``Person'' and ``Social Organization''
and
the simple description function ``age'' from the category ``Person'' to the datatype of natural numbers $\aleph$.
Define the crisp conceptual scale attribute ``youth organization'' by the query 
``What social organizations have only young members?''
This can be expressed mathematically as
$s \in \mbox{``Social Organization''}$
and for all $p \in \mbox{``Person''}$
we have
$\mbox{member}(p,s) \;\implication\; \mbox{age}(p,\mbox{``young''})$.
Using residuation,
this can be expressed as
$(s,\mbox{``young''}) \in (\psi \tensorimplysource (\phi_\ast \circ \sigma^{\rm op}))$,
where
\begin{scriptsizeflushleft}
$\begin{array}{r@{\hspace{2mm}=\hspace{2mm}}l}
{\cal G}^\prime                                             & {\tt Social}\;{\tt Organization}   \\
{\cal M} \stackrel{\sigma}{\rightharpoondown} {\cal D}      & \mbox{simple age conceptual scale} \\
{\cal G} \stackrel{\psi}{\rightharpoondown} {\cal G}^\prime & \mbox{membership relation}         \\
\psi \tensorimplysource (\phi_\ast \circ \sigma^{\rm op}) \type {\cal G}^\prime \rightharpoondown {\cal M} & \mbox{resulting facet}
\end{array}$
\end{scriptsizeflushleft}

This works quite well in the case of composite description functions consisting of simple description functions and binary relations,
and
many examples of composite description functions can be modeled using relational composition and residuation.
However,
it becomes more complicated in the case of n-ary relations.
Consider the context of corporate information.
Specifically,
information about the revenue rank of corporations
on a scale 0 to 100.
For investors,
an interesting query is: ``Has the revenue rank of the corporation been universally high recently?''
The processing here involves multiple ordinal scales
--- a Date ordinal scale and an ordinal scale on the subrange $[0,100]$ of natural numbers.
The metadata here is principally concerned with the ternary ``revenue rank'' relation
$\rho \subseteq \triproduct{{\cal C}}{{\cal D}}{[0,100]}$,
where 
${\cal C}$ represents the category of all corporations, and
${\cal D}$ is the date datatype.
Here we need to define the two attributes:
``high'' within a suitable conceptual scale on values $[0,100]$,
and 
``recently'' within a suitable conceptual scale on dates ${\cal D}$.
For example,
crisp specifications might be that
``high'' means any rank above 80
and
``recently'' means within the last 10 years.
Whatever the interpretations for these two terms,
the central question is how to handle the ternary relation $\rho$.
One method would be to use the process of reification,
which is used for interoperability between metadata standards \cite{kent98}.
Mathematically,
the relation $\rho$ is replaced by its three projection functions.
Even after this conversion from the n-ary to the binary case,
there yet remains the question of how to combine the scales with the metadata.
There appear to be several answers to this question.


\section{Summary and Future Work}

Conceptual scales are now being used for 
the interpretation, classification and organization of networked information resources \cite{kent98}.
An enriched notion of conceptual scale would provide for 
a more flexible approach in the interpretation of networked information.
Enriched conceptual scales equivalent to enriched linguistic variables
unify ideas from formal concept analysis, rough set theory, and fuzzy set theory.

Future work could possibly include any of the following initiatives.
The basic theorem \cite{wille82} of formal concept analysis should be developed in the enriched context.
Also in the area of formal concept analysis,
the basic operations on conceptual scales,
such as sums, products, and the important apposition operation,
should be further developed in the enriched setting.
It may be profitable to investigate possible connections, 
such as fixpoints, 
between enriched concept spaces and generalized metric spaces \cite{lawvere73,kent87}.
From the standpoint of soft computation,
there needs to be a closer integration of the enriched notions discussed in this paper
with the rough notions of formal concept analysis,
such as the rough formal concept \cite{kent93c}.
There should also be an integration of soft computation ideas,
such as discussed in this paper and elsewhere,
into current standards (Resource Description Framework, Conceptual Knowledge Markup Language, Conceptual Graph Markup Language, Knowledge Interchange Format and Conceptual Graph Interchange Format)
for conceptual knowledge representation and ontological modeling \cite{kent98}.



\begin{thebibliography}{10}

\bibitem{chen:etal94}
H.~Chen, P.~Hsu, R.~Orwig, L.~Hoopes, and J.F. Nunamaker.
\newblock Automatic concept classification of text from electronic meetings.
\newblock {\em Communications of the ACM}, 37:56--73, October 1994.

\bibitem{ganter89}
B.~Ganter and R.~Wille.
\newblock Conceptual scaling.
\newblock In F.~Roberts, editor, {\em Applications of Combinatorics and Graph
  Theory in the Biological and Social Sciences}, pages 139--167. Springer, New
  York, 1989.
\newblock Preprint No. 1174 (1988), Technische Hochschule Darmstadt, Darmstadt,
  Germany.

\bibitem{kent87}
R.E. Kent.
\newblock The metric closure powerspace construction.
\newblock In M.~Main, A.~Melton, M.~Mislove, and D.~Schmidt, editors, {\em
  Mathematical Foundations of Programming Semantics, 3rd Workshop}, pages
  173--199. Tulane University, New Orleans, Springer-Verlag, 1987.
\newblock Lecture Notes in Computer Science, Vol. 298.

\bibitem{kent92}
R.E. Kent.
\newblock Conceptual collectives.
\newblock Technical report, Department of Computer and Information Science,
  University of Arkansas at Little Rock, 1992.

\bibitem{kent94}
R.E. Kent.
\newblock Enriched interpretation.
\newblock In {\em Third International Workshop on Rough Sets and Soft Computing
  (RSSC'94)}, San Jose, California, USA, 1994.

\bibitem{kent93}
R.E. Kent.
\newblock Rough concept analysis.
\newblock In W.~Ziarko, editor, {\em Rough Sets, Fuzzy Sets and Knowledge
  Discovery}, pages 248--255. Springer-Verlag, August 1994.

\bibitem{kent93c}
R.E. Kent.
\newblock Rough concept analysis: A synthesis of rough sets and formal concept
  analysis.
\newblock {\em Fundamenta Informatica}, 27(2,3):169--181, August 1996.

\bibitem{kent98}
R.E. Kent.
\newblock Organizing conceptual knowledge online: Metadata interoperability and
  faceted classification.
\newblock In {\em Fifth International Conference of the International Society
  of Knowledge Organization (ISKO'98)}, Lille, France, August 1998.
\newblock To be distributed in the forthcoming issue of Knowledge Organization.

\bibitem{kent:neuss:94}
R.E. Kent and C.~Neuss.
\newblock Creating a web analysis and visualization environment.
\newblock {\em Computer Networks and ISDN Systems}, 28:107--117, 1995.

\bibitem{lawvere73}
F.W. Lawvere.
\newblock Metric spaces, generalized logic, and closed categories.
\newblock In {\em Seminario Mathematico E. Fisico}, volume~43, pages 135--166.
  Rendiconti, Milan, 1973.

\bibitem{pagliani96}
P.~Pagliani.
\newblock A modal relation algebra for generalized approximation spaces.
\newblock In {\em Rough Sets, Fuzzy Sets and Machine Discovery}, 1996.

\bibitem{pagliani97}
P.~Pagliani.
\newblock Information gaps as communication needs: A new semantic foundation
  for some non-classical logics.
\newblock {\em Journal of Logic, Language, and Information}, 6:63--99, 1997.

\bibitem{pawlak82}
Z.~Pawlak.
\newblock Rough sets.
\newblock {\em International Journal of Information and Computer Science},
  11:341--356, 1982.

\bibitem{rowley87}
J.~Rowley.
\newblock {\em Organising Knowledge: An Introduction to Information Retrieval}.
\newblock Gower Publishing Company Ltd., Hants, England, 1987.

\bibitem{wille82}
R.~Wille.
\newblock Restructuring lattice theory: An approach based on hierarchies of
  concepts.
\newblock In I.~Rival, editor, {\em Ordered Sets}, pages 445--470. Reidel,
  Dordrecht-Boston, 1982.

\bibitem{woods91}
W.A. Woods.
\newblock Understanding subsumption and taxonomy: A framework for progress.
\newblock In J.~Sowa, editor, {\em Principles of Semantic Networks:
  Explorations in the Representation of Knowledge}, pages 45--94. Morgan
  Kaufmann, San Mateo, California, 1991.

\bibitem{zadeh75}
L.~Zadeh.
\newblock The concept of a linguistic variable and its application to
  approximate reasoning.
\newblock {\em Information Science}, 8:199--249, 1975.

\end{thebibliography}

\end{document}